\documentclass[letterpaper]{article} % DO NOT CHANGE THIS
\usepackage{aaai25}  % DO NOT CHANGE THIS
\usepackage{times}  % DO NOT CHANGE THIS
\usepackage{helvet}  % DO NOT CHANGE THIS
\usepackage{courier}  % DO NOT CHANGE THIS
\usepackage[hyphens]{url}  % DO NOT CHANGE THIS
\usepackage{graphicx} % DO NOT CHANGE THIS
\usepackage{natbib}  % DO NOT CHANGE THIS AND DO NOT ADD ANY OPTIONS TO IT
\usepackage{caption} % DO NOT CHANGE THIS AND DO NOT ADD ANY OPTIONS TO IT
\usepackage{algorithm}
\usepackage{algorithmic}

\usepackage{newfloat}
\usepackage{listings}
\DeclareCaptionStyle{ruled}{labelfont=normalfont,labelsep=colon,strut=off} % DO NOT CHANGE THIS
\floatstyle{ruled}
\newfloat{listing}{tb}{lst}{}
\floatname{listing}{Listing}
\usepackage{booktabs}
\usepackage{xcolor,colortbl}
\usepackage{multirow}
\usepackage{amsmath}
\usepackage{amssymb}
\usepackage{siunitx}
\usepackage{enumitem}
\usepackage{pifont}
\usepackage[capitalize]{cleveref}
\usepackage{makecell}
\usepackage{diagbox}
\usepackage{array}

\renewrobustcmd{\boldmath}{}
\newrobustcmd{\B}{\bfseries}   
\newrobustcmd{\U}{\underline}
\newcommand{\cmark}{\ding{51}}%
\newcommand{\ra}[1]{\renewcommand{\arraystretch}{#1}}
\DeclareMathOperator*{\argminA}{arg\,min}

\title{Multi-View Pedestrian Occupancy Prediction with a Novel Synthetic Dataset}
\author{
    Sithu Aung\textsuperscript{\rm 1},
    Min-Cheol Sagong\textsuperscript{\rm 1},
    Junghyun Cho\textsuperscript{\rm 1,2,3}
}
\affiliations{
    \textsuperscript{\rm 1}Korea Institute of Science and Technology \
    \textsuperscript{\rm 2}AI-Robotics, KIST School, University of Science and Technology\\
    \textsuperscript{\rm 3} Yonsei-KIST Convergence Research Institute, Yonsei University \\
    \{sithu, mcsagong, jhcho\}@kist.re.kr
}

\begin{document}

\maketitle

\begin{abstract}
We address an advanced challenge of predicting pedestrian occupancy as an extension of multi-view pedestrian detection in urban traffic.
To support this, we have created a new synthetic dataset called \textbf{MVP-Occ}, designed for dense pedestrian scenarios in large-scale scenes.
Our dataset provides detailed representations of pedestrians using voxel structures, accompanied by rich semantic scene understanding labels, facilitating visual navigation and insights into pedestrian spatial information.
Furthermore, we present a robust baseline model, termed \textbf{OmniOcc}, capable of predicting both the voxel occupancy state and panoptic labels for the entire scene from multi-view images.
Through in-depth analysis, we identify and evaluate the key elements of our proposed model, highlighting their specific contributions and importance.
\end{abstract}

\begin{links}
    \link{Project page}{https://sithu31296.github.io/mvpocc}
\end{links}

\section{Introduction}
Recognition of crowded pedestrians in large-scale environments, particularly in surveillance systems with multiple CCTV cameras, has been an important area of research.
Various multi-view tasks including multi-view crowd counting \cite{mvms, cvcs}, multi-target multi-camera tracking \cite{mvmhat, jeon2023leveraging, luna2022graph, nguyen2023multi}, multi-view human pose estimation \cite{mvpose, voxelpose, mvp}, and multi-view pedestrian detection \cite{mvdet, shot, mvdetr, mvfp} have emerged in this area.
However, these tasks are still limited to the 2D ground plane when estimating crowd density and global positions of individuals. Although multi-view pose estimation methods yield 3D poses of pedestrians, they are yet constrained to small environments with a limited number of people and struggle with regressing metric-level 3D joint locations.

\begin{figure}[tb] \centering
    \includegraphics[width=\linewidth]{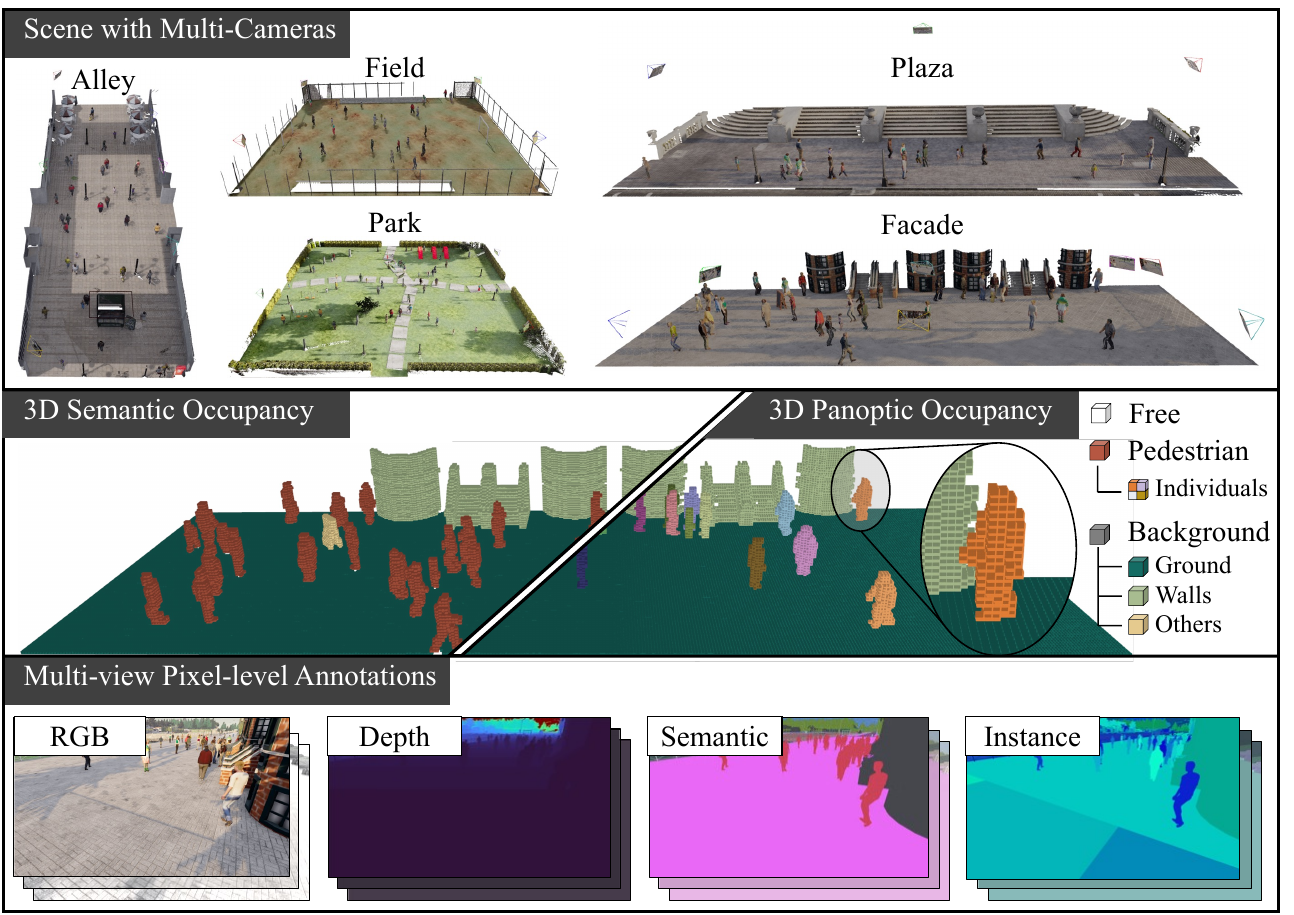}
    \caption{
    \textbf{Visualizations of the proposed dataset.}
    The primary objective is to predict the semantic and instance labels of the voxels and determine each pedestrian's location within the scene. The dataset includes five expansive scenes with dense pedestrian activity.
    (Best viewed in color.)
    }
    \label{fig:teaser}
\end{figure}

To address these limitations, we propose extending multi-view pedestrian detection to multi-view pedestrian occupancy prediction, as illustrated in \cref{fig:teaser}, where the occupancy status has to be predicted for every voxel in the entire scene and every grid cell within the bird's eye view plane.
Predicting occupancy status within the ground plane allows us to identify each pedestrian instance, while voxel-level occupancy prediction provides rich geometric information about the scene and a more articulate representation of pedestrians, including their heights.
Additionally, estimating the semantic label of each voxel enables us to gain insight into the surrounding contexts, which is an added advantage.
However, as highlighted in \cref{tab:ds_comparison}, existing datasets do not provide this level of information, underscoring the need for more comprehensive datasets tailored for spacious scenes with crowded pedestrians.

\begin{table}[tb]  \centering
  \setlength{\tabcolsep}{0.4mm}
  \begin{tabular}{l*{15}{c}}
    \toprule
    Dataset                 & Type   & Modality& \#frames& \#scenes & \#peds. & pose \\
    \midrule
    PETS2009                & Real   & I     & 0.42k & 1 & 40  &  - \\
    CMUPanoptic            & Real   & I+D   & 297k  & 1 & 8   & \cmark \\
    Human-M3                & Real   & I+P   & 12.2k & 3 & 10  & \cmark \\
    CityStreet              & Real   & I     & 0.5k  & 1 & 150 & - \\
    WildTrack               & Real   & I     & 0.4k  & 1 & 20  & - \\
    MultiviewX              & Syn. & I     & 0.4k  & 1 & 40  & - \\
    GMVD                    & Syn. & I     & 6k    & 7 & 40  & - \\
    \midrule
    \textbf{MVP-Occ}& Syn. & I+D+P & 12.5k & 5 & 100 & \cmark \\
  \bottomrule
  \end{tabular}
  \caption{
  \textbf{Comparison with available multiview pedestrian datasets.} Ours is by far the most comprehensive with rich 2D/3D annotations, including \textit{I}: image, \textit{D}: depth, \textit{P}: point cloud, human poses, segmentation, and occupancy labels.
  }
  \label{tab:ds_comparison}
\end{table}

Hence, we propose a novel synthetic \textbf{M}ulti-\textbf{V}iew \textbf{P}edestrian \textbf{Occ}upancy dataset, \textbf{MVP-Occ}, comprising five large-scale scenes, designed to mimic real-world environments.
In our dataset, the entire scene is represented by voxels, and each voxel is annotated with one of five classes, indicating whether it belongs to a pedestrian, the background environment, or is empty.
Furthermore, our dataset goes beyond occupancy prediction, offering a variety of perception tasks such as segmentation, pose estimation, semantic scene completion, depth estimation, and more, spanning both 2D and 3D domains.
To the best of our knowledge, our dataset is the first to provide such comprehensive 3D annotations for urban surveillance scenarios, opening avenues for further research in pedestrian recognition and scene understanding.

Alongside the newly proposed dataset, we introduce a robust baseline model, termed \textbf{OmniOcc} to tackle the challenge of multi-view pedestrian occupancy prediction.
Our model is characterized by its expandability and ability to handle various combinations of camera configurations and scene dimensions during train and test time.
Additionally, our model is designed to predict 2D pedestrian occupancy in the ground plane while simultaneously predicting 3D semantic occupancy for the entire scene.
Using pedestrian instances as center locations, our model can further group semantic occupancy into instance and panoptic occupancies.

Furthermore, we establish a new evaluation benchmark to assess the performance of the proposed model on both 2D and 3D occupancy predictions.
Our experiments cover conventional evaluations on the same scene and also address the challenging task of synthetic-to-real transfer with the WildTrack dataset~\cite{wildtrack}, for which we have generated the ground-truth segmentation data.
Through these extensive analyses, we conduct an in-depth examination of the proposed model, dissecting the individual contributions of each component.
The results underscore the superiority of our approach over previous multi-view detection methods, particularly highlighting its prowess in synthetic-to-real evaluation, where existing methods falter in transferring knowledge when confronted with disparate scenes.

\section{Related Work}
\subsubsection{Multi-view human pose estimation.}
Existing works \cite{mvpose, voxelpose, planesweeppose} on multi-view 3D pose estimation methods have predominantly concentrated on constrained environments with a limited number of individuals.
As summarized in \cref{tab:ds_comparison}, widely used datasets for this task, such as CMU-Panoptic \cite{cmupanoptic} and Human-M3 \cite{humanm3} are not intended for expansive scenarios involving a large number of pedestrians.
Consequently, there exists a demand for datasets that accurately portray complex CCTV-like situations, characterized by sparsely installed cameras that capture large scenes with dense pedestrian crowds.

\subsubsection{Multi-view pedestrian detection.}
Recent research efforts \cite{shot, mvdetr, 3drom, mvfp} have focused on detecting dense pedestrians in large scenes using multi-view information.
However, the datasets associated with these methods, such as WildTrack \cite{wildtrack}, MultiviewX \cite{mvdet}, and GMVD \cite{gmvd}, only provide pedestrian locations on the ground plane, lacking detailed information on poses and actions.
This limitation may restrict their utility in certain applications.
To address this gap and align with our research objectives, we aim to develop a dataset that includes detailed information about pedestrians and provides rich context at the scene level, thus facilitating a broader range of research applications.

\subsubsection{Semantic scene completion.}
Representing scenes with volumetric occupancy and semantic labels has been actively researched in the context of semantic scene completion (SSC) tasks.
SSC methods typically rely on pairs of RGB and depth images as input and require intermediate representations \cite{sscnet, torchssc, scfusion, tang2022not}.
Moreover, they focus mainly on small indoor environments, capturing static scenes and objects, while later advancements \cite{lmscnet, monoscene, ndcscene} expanded the SSC to outdoor scenes by inferring dense geometry and semantics from a single monocular image.
However, these methods still have limitations, including using a single RGB image as input and focusing on scene reconstruction with large objects.

\subsubsection{3D occupancy prediction.}
There has been a surge of interest in representing the surrounding environment around autonomous vehicles through dense occupancy and semantic labels using multiple cameras, following Tesla AI Day\cite{teslaaiday}.
Subsequent research works \cite{tpvformer, surroundocc, fbocc, occformer, sparseocc} have aimed to address this challenge through the 3D occupancy prediction task.
Concurrently, efforts \cite{occnet, occdata, openocc, occ3d} have been made to improve the quality of datasets by densifying the sparse occupancy labels obtained from the LiDAR sensors.
Although our proposed task shares similarities with the works mentioned earlier, there are distinct differences.
Our task focuses on scenarios with variable camera setups that capture static scenes with dense pedestrian crowds.
In contrast, above methods address scenarios involving a fixed camera setup installed on autonomous vehicles that observe dynamic scenes and prioritize large surrounding environments rather than individual pedestrians.

\section{Method}
\begin{figure*}[tb] \centering
  \includegraphics[width=\linewidth]{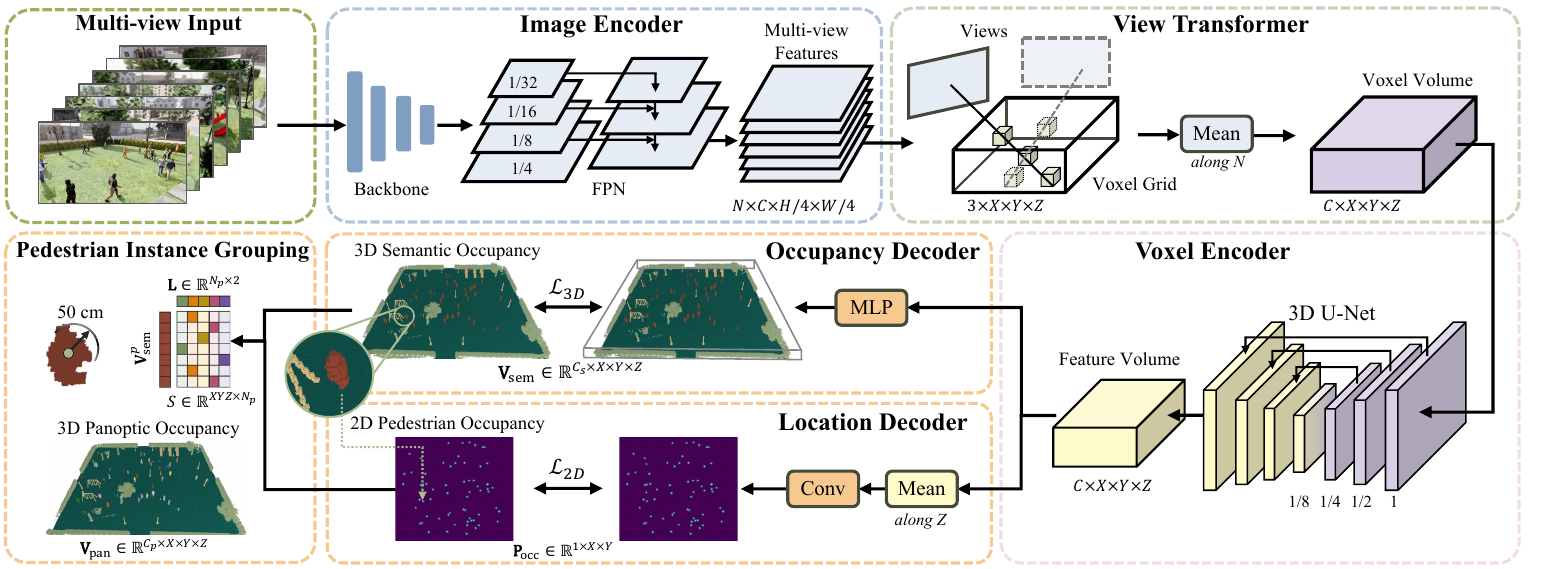}
  \caption{
  \textbf{Overview of the proposed model.}
  Image features are extracted using a backbone network augmented with an FPN.
  Next, multi-view 2D features are projected onto the voxel grid along rays and processed with a 3D U-Net to construct a feature volume.
  Semantic occupancy predictions are generated using a two-layer MLP network, whereas a single convolutional layer predicts the occupancy status of pedestrians.
  Finally, pedestrian instances are grouped on the basis of both predictions to obtain instance and panoptic occupancy labels.
  (Best viewed in color.)
  }
  \label{fig:model}
\end{figure*}

First, we introduce a baseline model to tackle the multi-view pedestrian occupancy prediction task.
This task entails predicting the location of each pedestrian and classifying voxels into one of the pre-defined semantic classes.
The subsequent sections provide detailed information about various components of the proposed model.

\subsection{Model Overview}
\cref{fig:model} shows an overview of the proposed model, which takes a set of $N$ posed multi-view images $\{I_n\}_{n=1}^{N}$.
Here, $I_n \in \mathbb{R}^{3\times H\times W}$ with $H$ height and $W$ width of the image.
The goal is to estimate dense 3D semantic, instance, and panoptic occupancies, represented by $\mathbf{V}_{\text{sem}} \in \mathbb{R}^{C_s\times X\times Y\times Z}$, $\mathbf{V}_{\text{pan}} \in \mathbb{R}^{C_p\times X\times Y\times Z}$, $\mathbf{V}_{\text{ins}} \in \mathbb{R}^{N_p\times X\times Y\times Z}$, respectively, along with 2D pedestrian occupancy $\mathbf{P}_{\text{occ}} \in \mathbb{R}^{1\times X\times Y}$.
In these expressions, $C_s$ and $C_p$ indicate the number of semantic and panoptic classes, respectively, $N_p$ denotes the total number of pedestrians within the scene, while $X$, $Y$, and $Z$ denote the dimensions of the voxel grid that defines the scene.

\subsubsection{Image encoder.} 
Given that the input modality is in pixel space, we employ a ResNet \cite{resnet} model to extract multi-scale 2D features.
These features are instrumental in capturing diverse semantic and contextual information across different scales, which is particularly advantageous for detecting small pedestrians within spacious scenes.
The multi-scale features are effectively fused with an FPN \cite{fpn} network.
The spatial resolution of the fused feature map is set to one-fourth of the original image size.

\subsubsection{View transformer.} 
In estimating the occupancy status and semantic properties of the entire scene, a crucial step involves translating the multi-view features onto the 3D space.
Recent query-based feature transformation methods\cite{tpvformer, sparseocc, surroundocc} are not suitable for our context due to various camera configurations and distinct scene dimensions.
As an alternative, we adopt a non-parametric feature transformation approach \cite{imvoxelnet}, which has been shown to achieve comparable performance \cite{simplebev}.
This method operates under the assumption of a uniform depth distribution along the ray, implying that all voxels along a camera ray are filled with identical features corresponding to a single pixel in a 2D space.

To implement this process, we first construct a voxel grid $V_n \in \mathbb{R}^{3\times X\times Y\times Z}$ that contains voxel centers and corresponds to the $n$-th view. 
The value of each center of the voxel $\mathbf{p} = (x, y, z)^T$ is interpolated bilinearly from the image feature $F_n$ at the projected pixel coordinates $(u_n, v_n)$ of $\mathbf{p}$.
Afterward, these feature samples are averaged from all views, considering only the valid number of 2D projections as \cref{eq:view_transformation,eq:view_transformation_m,eq:view_transformation_v}.

\begin{equation}
\label{eq:view_transformation}
    \mathbf{V}(\mathbf{p}) = \frac{1}{N}\sum_{n=1}^N M_n(\mathbf{p}) V_n(\mathbf{p}),
\end{equation}

\begin{equation}
\label{eq:view_transformation_m}
     M_n(\mathbf{p}) = 
    \begin{cases}
        1 & \text{if} \; 0 \leq u_n \le \frac{W}{4} \; \text{and} \; 0 \leq v_n \le \frac{H}{4},\\
        0 & \text{otherwise},
    \end{cases}
\end{equation}

\begin{equation}
\label{eq:view_transformation_v}
    V_n(\mathbf{p})=\operatorname{interp}((u_n, v_n), F_n),
\end{equation}
where $(u_n, v_n)^T = K_n' R_n (p, 1)^T$, $K_n'$ is the scaled intrinsic matrix, $R_n$ is the extrinsic matrix, and $M_n$ is a binary mask indicating whether the projected $\mathbf{p}$ is inside the camera's frustum or not.

\subsubsection{Voxel encoder.} 
A 3D U-Net \cite{unet3d} based network is applied to process the voxel volume leveraging its effectiveness in handling volumetric data across various 3D tasks \cite{simpleocc, occdepth}.
It consists of a stack of three residual blocks, each with three downsampling 3D convolutional layers followed by three transposed 3D convolutional layers with $C=64$ channels.
This allows the network to capture and refine hierarchical features within the voxel volume, enabling precise occupancy prediction and semantic label assignment.
The channel dimension is doubled in each downsampling layer and halved in each upsampling layer, following \cite{atlas}.

\subsubsection{Occupancy decoder.} 
As the voxel encoder already has a sufficient learnable capacity to accurately infer geometric and semantic information, we opt for a straightforward design for the 3D occupancy decoder.
We utilize a two-layer MLP network to produce semantic occupancy prediction $\mathbf{V}_{\text{sem}}$, encapsulating the classification of each voxel into one of five distinct semantic classes ($C_s = 5$).

\subsubsection{Location decoder.} 
We compress the feature volume by averaging along the vertical dimension ($Z$), resulting in a bird's eye view (BEV) feature map.
A single convolutional layer is then applied to generate a 2D occupancy map $\mathbf{P}_{\text{occ}}$, which indicates the presence or absence of pedestrians within each cell of the BEV grid.
We find that additional processing of the feature volume does not enhance location accuracy.
This suggests that the pedestrians' locations are primarily inferred from the occupancy status of 3D voxels.

\subsubsection{Pedestrian instance grouping.} 
Taking advantage of the simultaneous prediction of $\mathbf{V}_{\text{sem}}$ and $\mathbf{P}_{\text{occ}}$, we group the semantic occupancy of pedestrian class, $\mathbf{V}_{\text{sem}}^p$ into $\mathbf{V}_{\text{ins}}$.
Let $\mathbf{L} \in \mathbb{R}^{N_p \times 2} = \{\mathbf{l}_j\}_{j=1}^{N_p}$ denote the set of $N_p$ pedestrian locations, obtained by filtering $\mathbf{P}_{\text{occ}}$ with a pre-defined confidence threshold $\tau$ and applying non-maximum suppression.
An affinity matrix $S \in \mathbb{R}^{XYZ \times N_p}$ can be defined where each element $S_{i,j}$ represents the euclidean distance between the $i$-th voxel and $j$-th location:

\begin{equation}
    S_{i,j} = \|\mathbf{v}_i - \mathbf{l}_j\|,
\end{equation}

\begin{equation}
    W_{i,j} = 
    \begin{cases}
        1 & \text{if} \: S_{i,j} < r,\\
        0 & \text{otherwise},
    \end{cases}
\end{equation}
where $\mathbf{v}_i$ is the coordinate vector with $x$ and $y$ coordinates of $\mathbf{p}$.
$W_{i,j}$ is a binary mask to check if $S_{i,j}$ is within a threshold $r$ of \SI{50}{\centi\metre}. 
Then, for each voxel $i$, we can find the index of the closest location $j$:

\begin{equation}
    D_i = \argminA_j\: S_{i,:N_p},
\end{equation}
and if $\exists i$ such that $W_{i,j} = 1$, we set the label of $i$-th voxel in $\mathbf{V}_{\text{ins}}$ to $D_i$.
We obtain panoptic occupancy prediction $\mathbf{V}_{\text{pan}} \in \mathbb{R}^{C_p\times X\times Y\times Z}$ after combining $\mathbf{V}_{\text{ins}} \in \mathbb{R}^{N_p\times X\times Y\times Z}$ with three background/stuff classes of $\mathbf{V}_{\text{sem}}$, where $C_p = C_b + N_p$ and $C_b = 3$.

\subsection{Loss}
\label{sec:loss}

\subsubsection{3D occupancy loss.} 
Discretizing the vast expanse of the scene results in a large amount of free voxels.
To address the dominance of the Free class and ensure balanced optimization across all classes, we use a weighted cross-entropy loss, $\mathcal{L}_{\text{wce}}$.
The weights are derived from the inverse of the class frequency $f_c$, computed as $w = \frac{1}{\log(f_c+\epsilon)}$, where $\epsilon \ll 1$, following \cite{lmscnet, rangenet++}.
Furthermore, we incorporate the Lovász-Softmax loss $\mathcal{L}_{\text{lovasz}}$ \cite{lovasz} to enhance segmentation quality for pedestrians.
We also apply the scene-class affinity loss $\mathcal{L}_{\text{affinity}}$ \cite{monoscene} to further improve overall segmentation quality by optimizing geometry and semantics separately.
The final 3D occupancy loss is formulated as follows:

\begin{equation}
    \mathcal{L}_{3D} = \lambda_{\text{wce}} \cdot \mathcal{L}_{\text{wce}} + \lambda_{\text{lovasz}} \cdot \mathcal{L}_{\text{lovasz}} + \lambda_{\text{affinity}} \cdot \mathcal{L}_{\text{affinity}},
\end{equation}
where $\lambda_{\text{wce}} = 0.4$, $\lambda_{\text{lovasz}} = 0.3$, and $\lambda_{\text{affinity}} = 0.3$ are hyperparameters to balance the loss components.

\subsubsection{2D occupancy loss.}
Occupancy prediction in the BEV plane resembles a keypoint detection problem, aiming to produce a heatmap that reflects the likelihood of occupancy at each position on the ground plane.
We utilize a Gaussian kernel to splat all ground-truth pedestrian locations onto a heatmap and compute the mean squared error (MSE) between the predicted and ground-truth occupancy maps.

\subsubsection{Total loss.}
The final loss function is a composite of both 3D and 2D occupancy losses, expressed as:
\begin{equation}
    \mathcal{L} = (1 - \lambda) \cdot \mathcal{L}_{3D} + \lambda \cdot \mathcal{L}_{2D},
\end{equation}
where $\lambda = 0.3$ is a weighting coefficient and prioritizes $\mathcal{L}_{3D}$ due to its higher complexity and difficulty in optimization.

\section{Datasets}
\begin{table}[tb]  \centering
  \setlength{\tabcolsep}{0.8mm}
  \begin{tabular}{l|*{5}{c}|c}
    \toprule
    Scenes     & Alley & Plaza & Field & Park & Facade & WildTrack \\
    \midrule
    \small{\#cameras}     & 6 & 3 & 4 & 8 & 7 & 7 \\
    \small{size(m)}     & \small{18$\times$45} & \small{15$\times$46} & \small{29$\times$40} & \small{48$\times$43} & \small{36$\times$12} & \small{12$\times$36} \\
    \small{\#peds.} & 60 & 50 & 40 & 100 & 40 & 20\\
  \bottomrule
  \end{tabular}
  \caption{
  \textbf{Comparison between MVP-Occ and WildTrack.}}
  \label{tab:scene_comp}
\end{table}

\begin{table*}[t] \centering
  \setlength{\tabcolsep}{0.85mm}
  \setlength{\aboverulesep}{0pt}
  \setlength{\belowrulesep}{0pt}
  \ra{1.1}
  \begin{tabular}{lll|ccc|ccc|ccc|ccc|ccc}
    \toprule
    \diagbox[width=2\tabcolsep]{}&\multicolumn{2}{l|}{\small{Method}}& \multicolumn{3}{c|}{MVDet$^*$} & \multicolumn{3}{c}{SHOT$^*$} & \multicolumn{3}{|c}{GMVD$^\dagger$} & \multicolumn{3}{|c}{MVFP$^\dagger$} & \multicolumn{3}{|c}{OmniOcc (Ours)$^\dagger$}\\ 
    \cmidrule{2-2} \cmidrule(lr){4-6} \cmidrule(lr){7-9} \cmidrule(lr){10-12} \cmidrule(lr){13-15} \cmidrule(lr){16-18}
    \multicolumn{2}{l}{\small{Training}}&\diagbox[width=2\tabcolsep]{}& MODA & MODP & F1 & MODA & MODP & F1 & MODA & MODP & F1 & MODA & MODP & F1 & MODA & MODP & F1 \\
    \midrule
    \multicolumn{3}{l|}{Alley}   & 93.4 & 87.9 & 96.6 & 95.0 & \B 91.0 & 97.5 & 91.9 & 88.3 & 95.8 & \U{95.8} & \U{90.8} & \U{97.8} & \B 96.8 & 88.9 & \B 98.0 \\
    \multicolumn{3}{l|}{Plaza}   & 87.5 & 88.5 & 93.3 & 88.9 & \U{89.5} & 94.2 & 85.5 & 87.4 & 92.3 & \U{89.3} & \B 90.7 & \U{94.4} & \B 92.4 & 86.7 & \B 96.1 \\
    \multicolumn{3}{l|}{Field}   & 82.1 & 85.6 & 90.5 & \U{85.3} & \B 88.9 & \U{92.2} & 82.4 & 85.2 & 90.6 & 84.9 & \U{87.1} & \U{92.2} & \B 92.6 & 86.9 & \B 96.3 \\
    \multicolumn{3}{l|}{Park}    & 88.9 & 84.9 & 94.2 & 88.5 & 87.3 & 93.9 & 88.4 & 84.5 & 93.9 & \U{91.2} & \U{87.5} & \U{95.5} & \B 93.4 & \B 87.9 & \B 96.6 \\
    \multicolumn{3}{l|}{Facade}  & 91.4 & 87.4 & 95.5 & 92.0 & 89.5 & 95.8 & 90.1 & 88.1 & 94.8 & \U{92.7} & \B 90.9 & \U{96.2} & \B 93.7 & \U{89.8} & \B 96.8 \\
    \midrule
    \multicolumn{3}{l|}{Avg.}    & 88.7 & 86.7 & 94.0 & 89.9 & \U{89.2} & 94.7 & 87.7 & 86.7 & 93.5 & \U{90.8} & \B 89.4 & \U{95.2} & \B 93.8 & 88.0 & \B 96.8 \\
  \bottomrule
  \end{tabular}
  \caption{
  \textbf{2D pedestrian occupancy prediction under same-scene evaluation on MVP-Occ.}
  Previous methods are supervised with ground-plane locations, while our method additionally utilizes voxel-level supervision.
  $^*$ methods are scene-specific, while $^\dagger$ methods can work with variable camera setups.
  The best and second best results are \textbf{bolded}, and \U{underlined}, respectively.
  }
  \label{tab:2d_occ}
\end{table*}

This section presents MVP-Occ, our new synthetic dataset, and describes how we added new labels to the existing real-world WildTrack dataset \cite{wildtrack} for evaluation purposes.

\subsection{MVP-Occ Dataset}
Our dataset is generated using the CARLA simulator \cite{carla}, originally developed for autonomous driving research \cite{shift, aiodrive, drivelm, carff}. 
Due to its comprehensive functionalities and rapid development capabilities, we have adapted it to suit our urban traffic scenario.

\subsubsection{Scene generation and simulation.}
CARLA offers diverse environments, from which we select specific scenes for our study.
Similarly to Human-M3\cite{humanm3} and GMVD\cite{gmvd}, we provide multiple scenes to investigate cross-scene performance, named Alley, Plaza, Field, Park and Facade, as described in \cref{tab:scene_comp}.
Each scene is designed with unique characteristics to challenge the model's adaptability and robustness.
Pedestrian models from the built-in assets library are utilized to simulate foot traffic, including activities such as walking and running.
Pedestrians are strategically spawned within and outside the scene to emulate entry and exit patterns and navigate through the scene through AI control, which enables obstacle avoidance and autonomously generates navigation paths to predefined target locations.
In addition, manual intervention is applied to manipulate certain pedestrians, ensuring their presence within or departure from the scene upon reaching their designated destinations.
Moreover, we control the timing and speed of certain pedestrians to simulate group motions, such as synchronized walking or following behaviors, and maintaining flock formation.

\subsubsection{Rendering and data generation.}
We render a comprehensive suite of sensors, including RGB, depth, semantic, and instance segmentation sensors, as depicted in \cref{fig:teaser}.
The camera setup configuration and the degree of overlapping fields of view are meticulously adjusted for each scene.
The dataset is generated at 10 FPS for 2500 frames, with an image size of 1920$\times$1080.
Point clouds obtained from monocular views are fused and subsequently voxelized within the area of interest to get occupancy labels.
Details are provided in the supplementary material.
A \SI{10}{\centi\metre} voxel size is used for discretization and is categorized into one of five unified classes: Free, Pedestrian, Ground, Wall, and Others with a supercategory of Background class for all scenes.
Note that in our dataset, only one instance/thing class (Pedestrian) exists with three stuff classes (Ground, Wall, and Others) for the panoptic occupancy prediction task.

\subsection{WildTrack Dataset}
WildTrack \cite{wildtrack} is a real-world dataset captured using seven cameras with significant overlapped fields of view.
We aim to evaluate synthetic-to-real performance on this dataset by training on the proposed synthetic scenes.
However, the dataset only provides pedestrian locations on the ground plane and lacks occupancy labels, which are difficult to obtain in real-world scenes.
To address this, we assess 3D occupancy prediction performance by comparing the rendered occupancy predictions with 2D segmentation masks.
Due to the absence of segmentation data in WildTrack, we manually annotated each image, generating 2D semantic, instance, and panoptic segmentation labels.
This approach enables us to validate our model's performance in predicting accurate 2D segmentations, serving as a proxy for its 3D occupancy prediction capabilities.

\begin{table}[t] \centering
  \setlength{\tabcolsep}{0.8mm}
  \setlength{\aboverulesep}{0pt}
  \setlength{\belowrulesep}{0pt}
  \ra{1.1}
  \begin{tabular}{lll|*{3}{c}|*{1}{c}|*{3}{c}}  
    \toprule
    \diagbox[width=2\tabcolsep]{}&\multicolumn{2}{l}{\small{Metric}} & \multicolumn{3}{|c}{Semantic Occ.} & \multicolumn{1}{|c}{Inst Occ.} & \multicolumn{3}{|c}{Panoptic Occ.}\\ 
    \cmidrule{2-2} \cmidrule(lr){4-6} \cmidrule(lr){7-7} \cmidrule(lr){8-10}
    \multicolumn{2}{l}{\small{Training}}&\diagbox[width=2\tabcolsep]{}& mIoU & IoU\textsubscript{Ped.} & IoU\textsubscript{Bg.} & AP & PQ & SQ & RQ \\
    \midrule
    \multicolumn{3}{l|}{Alley}   & 93.6 & 68.5 & 99.9 & 95.8 & 95.3 & 96.0 & 99.1\\
    \multicolumn{3}{l|}{Plaza}   & 93.5 & 67.6 & 99.9 & 92.0 & 94.6 & 96.1 & 98.2\\
    \multicolumn{3}{l|}{Field}   & 93.6 & 68.2 & 99.9 & 93.6 & 94.5 & 96.5 & 97.7\\
    \multicolumn{3}{l|}{Park}    & 91.8 & 62.9 & 98.9 & 95.5 & 94.5 & 95.9 & 98.3\\
    \multicolumn{3}{l|}{Facade}  & 93.7 & 69.5 & 99.8 & 94.8 & 95.9 & 97.0 & 98.8\\
  \bottomrule
  \end{tabular}
  \caption{
  \textbf{3D occupancy prediction under same-scene evaluation on MVP-Occ.}
  }
  \label{tab:3d_occ}
\end{table}

\begin{figure}[tb] \centering
  \includegraphics[width=\linewidth]{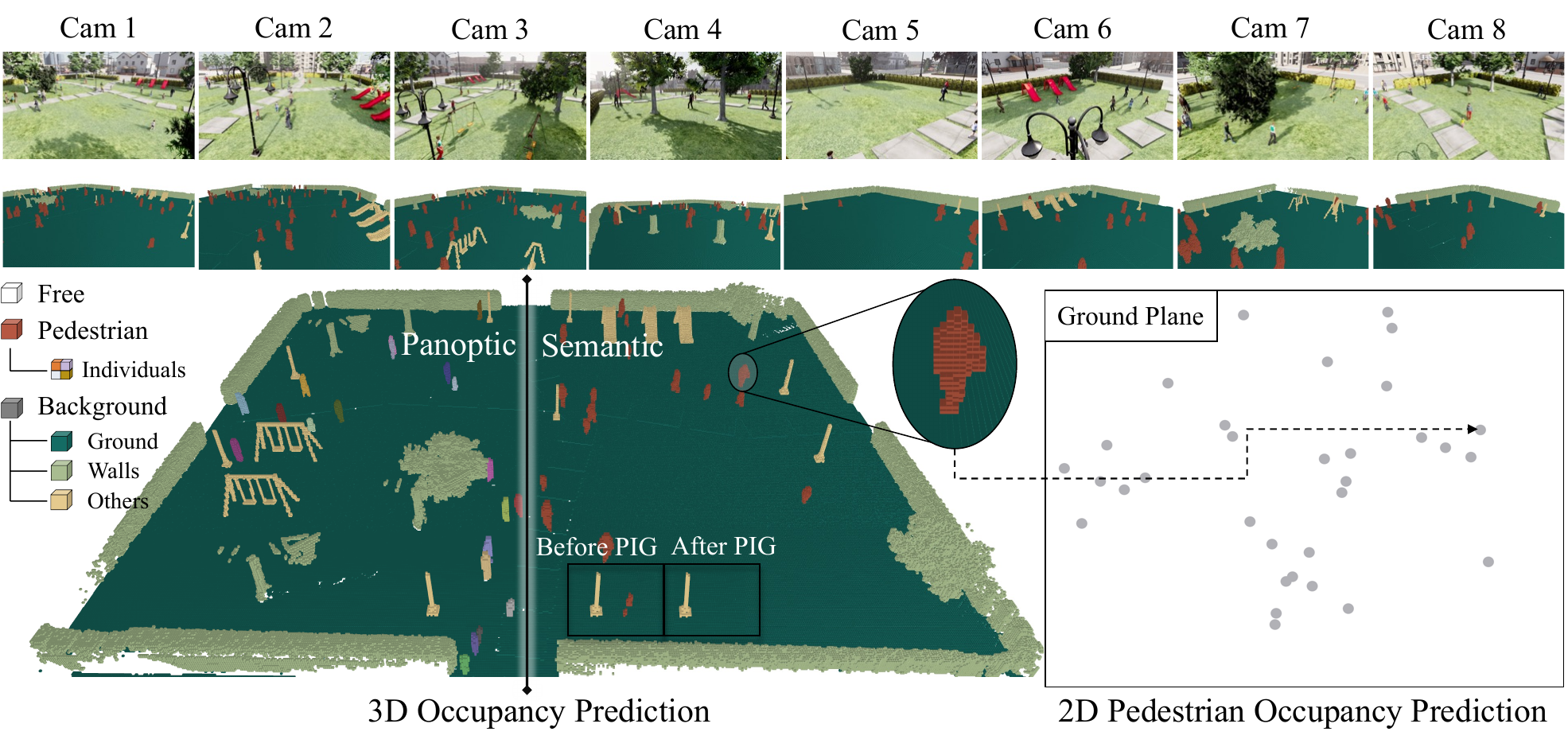}
  \caption{
  \textbf{Qualitative results of 2D and 3D occupancy predictions under same-scene evaluation on the Park scene.}
  (Best viewed in color.)
  }
  \label{fig:same_scene}
\end{figure}

\section{Experiments}
\begin{table*}[t] \centering
\begin{minipage}[t]{0.44\linewidth} \centering
    \setlength{\tabcolsep}{0.75mm}
    \setlength{\aboverulesep}{0pt}
    \setlength{\belowrulesep}{0pt}
    \ra{1.1} % Adjust the row height here (1.5 is 50% more height)
    \begin{tabular}{lll|*{2}{c}|*{2}{c}|*{2}{c}}
    \toprule
    \diagbox[width=2\tabcolsep]{}&\multicolumn{2}{l|}{\small{Method}} & \multicolumn{2}{c|}{GMVD} & \multicolumn{2}{c|}{MVFP} & \multicolumn{2}{c}{OmniOcc \small{(Ours)}}\\ 
    \cmidrule{2-2} \cmidrule(lr){4-5} \cmidrule(lr){6-7} \cmidrule(lr){8-9}
    \multicolumn{2}{l}{\small{Training}}&\diagbox[width=2\tabcolsep]{}& MODA & F1 & MODA & F1 & MODA & F1 \\
    \midrule
    \multicolumn{3}{l|}{Alley} & 11.8 & 34.1 & \U{11.9} & \U{36.5} & \B 33.3 & \B 63.2 \\
    \multicolumn{3}{l|}{Plaza} &  7.5 & 15.5 & \U{14.3} & \U{35.1} & \B 15.0 & \B 54.1 \\    % finished
    \multicolumn{3}{l|}{Field} &  5.2 & 10.0 & \U{24.8} & \U{41.9} & \B 41.2 & \B 60.4 \\ 
    \multicolumn{3}{l|}{Park}  &  1.8 & 24.5 & \U{18.5} & \U{47.6} & \B 48.7 & \B 72.1 \\ 
    \multicolumn{3}{l|}{Facade}& 46.3 & 71.9 & \U{52.6} & \U{74.9} & \B 75.4 & \B 87.5 \\    % finished
   \bottomrule
   \end{tabular}
   \caption{
   \textbf{2D pedestrian occupancy prediction under synthetic-to-real evaluation on WildTrack. }
   Our model outperforms previous methods in all scenes.
   }
   \label{tab:2d_occ_cross}
  
\end{minipage}\hfill
\begin{minipage}[t]{0.53\linewidth} \centering
    \setlength{\tabcolsep}{0.75mm}
    \setlength{\aboverulesep}{0pt}
    \setlength{\belowrulesep}{0pt}
    \ra{1.1}
    \begin{tabular}{lll|*{3}{c}|*{3}{r}|*{3}{c}}
    \toprule
    \diagbox[width=2\tabcolsep]{}&\multicolumn{2}{l|}{\small{Metric}} & \multicolumn{3}{c|}{Semantic Occ.} & \multicolumn{3}{c}{Instance Occ.} & \multicolumn{3}{|c}{Panoptic Occ.} \\ 
    \cmidrule{2-2} \cmidrule(lr){4-6} \cmidrule(lr){7-9} \cmidrule(lr){10-12} 
    \multicolumn{2}{l}{\small{Training}}&\diagbox[width=2\tabcolsep]{}& mIoU & IoU\textsubscript{Ped.} & IoU\textsubscript{Grd.} & AP & AP\textsubscript{25} & AP\textsubscript{50} & PQ & SQ & RQ \\
    \midrule
    \multicolumn{3}{l|}{Alley}  & 74.5 & 46.1 & 89.5 & \U{25.8} & \U{55.0} & \U{21.2}  & \U{54.5} & \U{75.8} & \U{65.2} \\ 
    \multicolumn{3}{l|}{Plaza}  & 34.1 & 19.0 & 36.6 & 3.7 & 12.8 & 7.3  & 43.5 & 73.2 & 50.8 \\ 
    \multicolumn{3}{l|}{Field}  & 62.3 & 42.6 & 78.5 & 20.9 & 41.4 & 16.3  & 43.7 & 69.3 & 57.6 \\ 
    \multicolumn{3}{l|}{Park}   & \U{75.2} & \U{47.9} & 89.7 & 23.4 & 47.8 & 18.4  & 54.1 & 75.6 & 64.7 \\ 
    \multicolumn{3}{l|}{Facade} & \B 79.8 & \B 58.4 & \B 90.7 & \B 40.2 & \B 70.0 & \B 39.8  & \B 58.9 & \B 76.2 & \B 72.3 \\ 
    \bottomrule
    \end{tabular}
  \caption{
  \textbf{3D occupancy prediction under synthetic-to-real evaluation on WildTrack.}
  Note that the evaluation is done with multi-view segmentation data due to the lack of occupancy ground-truths.
  }
  \label{tab:3d_occ_cross}
\end{minipage}
\end{table*}

\begin{figure}[tb] \centering
  \includegraphics[width=\linewidth]{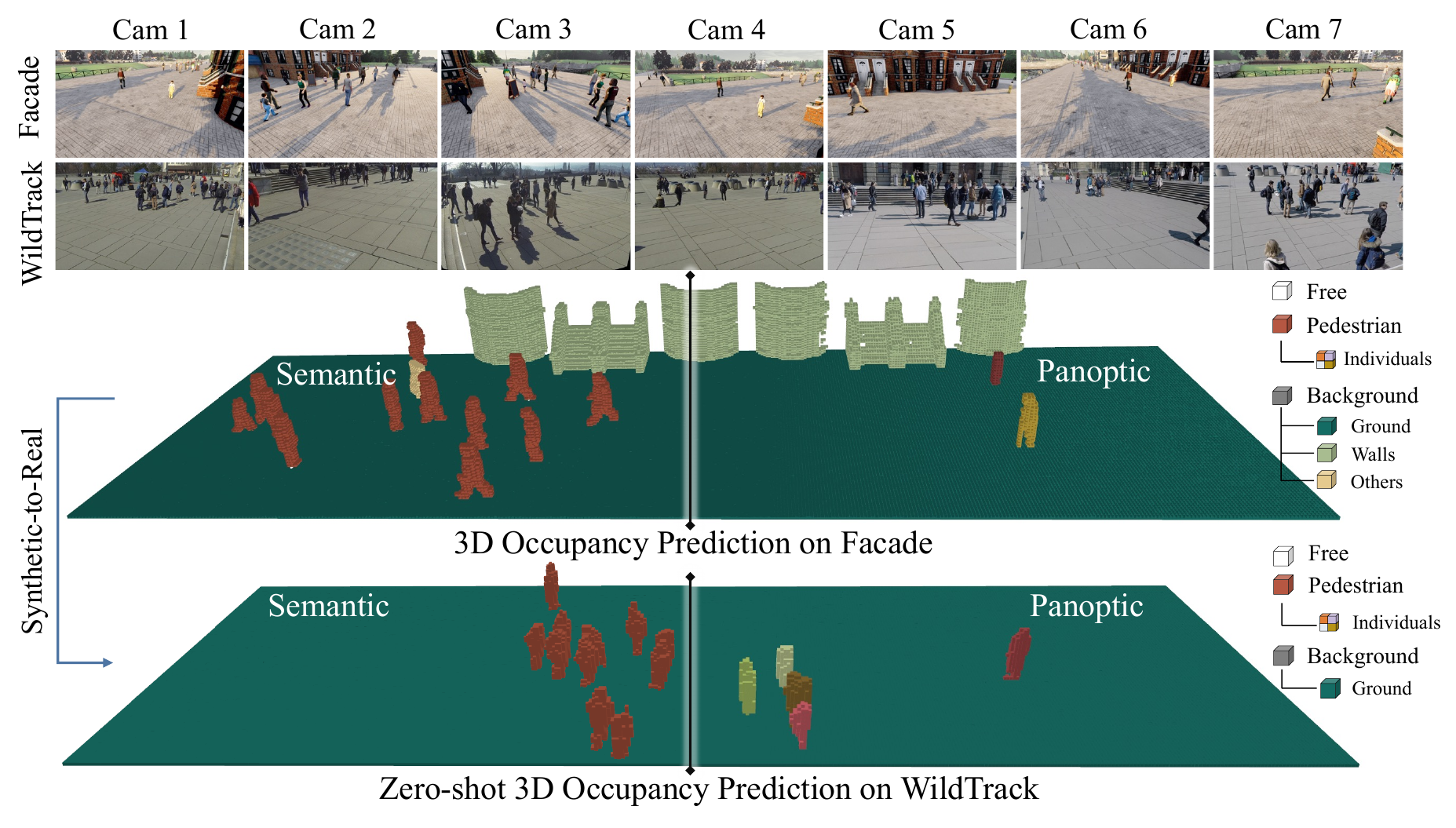}
  \caption{
  \textbf{Qualitative results of synthetic-to-real transfer from Facade to WildTrack.}
  (Best viewed in color.)
  }
  \label{fig:synctoreal}
\end{figure}

In this section, we first report on the evaluation metrics and training details. 
The subsequent sections provide extensive experiments on both 2D and 3D occupancy prediction benchmarks and lastly, ablation studies.

\subsubsection{Evaluation metrics.} 
We use commonly adopted metrics from multi-view detection methods to gauge the performance of 2D occupancy prediction.
These include MODA (Multi-Object Detection Accuracy), MODP (Multi-Object Detection Precision), and F1 score, each offering insights into the model's proficiency in pedestrian detection within the scene.
For 3D semantic occupancy, we calculate voxel-level IoU metrics to quantify the overlap between predicted and ground-truth occupancies for each semantic class and mean IoU across all classes.
We report AP (Average Precision) as mask AP using pycocotools~\cite{microsoftcoco} for 3D instance occupancy and PQ (Panoptic Quality) metric \cite{panopticapi}, which is composed of SQ (Segmentation Quality) and RQ (Recognition Quality) for 3D panoptic occupancy.
Details are provided in the supplementary material.

\subsubsection{Training details.}
Multi-view images are resized to 1280$\times$720, and ResNet-18 is used as the backbone network.
Each scene in our dataset is divided into training and testing sets, allocating 80\% of the initial frames to the training split and reserving the remaining 20\% for testing purposes, while WildTrack uses a 90/10 split.
A voxel size of \SI{10}{\centi\metre} for discretization results in variable grid dimensions, depending on the scene dimensions, as detailed in \cref{tab:scene_comp}.
$\tau$ is set to 0.5 to filter out low-confidence detections in 2D occupancy prediction.
The optimization process employs the AdamW optimizer with an initial learning rate of $1\times 10^{-3}$ and a decay rate of $1\times 10^{-2}$.
Training is conducted over 5 epochs, with a cosine learning rate scheduler dynamically adjusting the learning rate.
The experiments were performed using four NVIDIA A100 GPUs, with a batch size of 4.

\subsection{Same-Scene Evaluation}
In this evaluation scheme, all models undergo training in the train set and are evaluated on the test set of each scene in the proposed MVP-Occ dataset.

\subsubsection{2D occupancy prediction.}   
We compare our model with previous multi-view detection methods, MVDet~\cite{mvdet}, SHOT~\cite{shot}, GMVD~\cite{gmvd}, and MVFP~\cite{mvfp}, using the 2D occupancy prediction benchmark in \cref{tab:2d_occ}.
Our model outperforms all previous methods trained solely with 2D occupancy maps.
However, despite an overall improvement in detection, our model yields a lower MODP score in most scenes, which measures the precision of true positive instances relative to the ground truth.
The observed discrepancy can be attributed to the inherent misalignment between regressing each pedestrian's location on the ground plane and the voxel-level whole-body representation.

\subsubsection{3D occupancy prediction.}
In \cref{tab:3d_occ}, we present the same-scene performance of the proposed model on the new 3D occupancy prediction benchmark, which includes semantic, instance, and panoptic occupancies.
As anticipated, under the same-scene evaluation protocol, the model demonstrates exceptional performance, achieving nearly 99\% IoU for the Background class and 62-68\% IoU for the Pedestrian class, depending on the scene.
Qualitative results for the Park scene are provided in \cref{fig:same_scene}, showcasing the model's capability to effectively reconstruct the scene and accurately detect pedestrians when provided with ground-truth data for the scene of interest.

\subsection{Synthetic-to-Real Evaluation}
In this evaluation scheme, all models are trained on each synthetic scene of the proposed MVP-Occ dataset and then evaluated on the real WildTrack scene.

\subsubsection{2D occupancy prediction.}  
We compare against previous cross-scene multi-view detection models, GMVD~\cite{gmvd} and MVFP~\cite{mvfp} in \cref{tab:2d_occ_cross}.
Unlike the Facade scene, other scenes in our dataset exhibit significant differences from WildTrack, including variations in camera setup and scene characteristics.
Despite these challenging scenarios, the proposed model demonstrates superior performance across all scenes compared to previous methods, which often struggle with disparate scenes.
Notably, GMVD, which relies on ground plane detection, fails to transfer whereas MVFP, which utilizes 3D feature pulling, shows better generalization performance.
Our OmniOcc model surpasses both by incorporating scene understanding, which helps mitigate these issues.
The Park scene, with its diverse pedestrian distribution further enhances our model's ability to focus on pedestrians, resulting in improved pedestrian detection performance.
All models perform well when using the Facade scene, which was designed to closely resemble WildTrack's conditions.

\subsubsection{3D occupancy prediction.}
We justify the performance of the 3D occupancy prediction by evaluating the rendered 2D segmentation masks with manually labeled ground-truth 2D segmentation data.
Unlike the scenes in our dataset, the WildTrack dataset has only three semantic classes: Free, Pedestrian, and Ground.
Based on \cref{tab:3d_occ_cross}, the 3D semantic occupancy prediction yields conclusions similar to the 2D occupancy prediction.
However, for 3D instance and panoptic occupancy predictions, we observe better segmentation quality for individual pedestrians in the Alley scene compared to the Park scene.
The best performance is consistently achieved by the Facade scene, which continues to deliver outstanding results across all occupancy prediction tasks.
The qualitative results in \cref{fig:synctoreal} further demonstrate the impressive generalization capabilities of the Facade scene in WildTrack.
The segmentation quality of individual pedestrians is notably more accurate for the Facade scene, where the model was trained.
Nevertheless, addressing the challenges associated with scene understanding in dissimilar scenes requires further investigation and methodological refinements.

\begin{table}[t] \centering
\setlength{\tabcolsep}{1.3mm}
\begin{tabular}{*{10}{c}}
    \toprule
    $\mathcal{L}_{\text{wce}}$ & $\mathcal{L}_{\text{lovasz}}$ & $\mathcal{L}_{\text{affinity}}$ & mIoU & IoU\textsubscript{Ped.} & IoU\textsubscript{Others}\\
    \midrule
       -   &    -   &    -   & 85.3 & 67.1 & 59.9 \\     
    \cmark &    -   &    -   & 89.4 & 67.2 & 79.8 \\      
    \cmark & \cmark &    -   & \U{93.4} & \U{67.9} & \U{99.7} \\
    \cmark &    -   & \cmark & \U{93.4} & 67.4 & \B 99.9 \\
    \cmark & \cmark & \cmark & \B 93.7 & \B 69.5 & \U{99.7} \\
    \bottomrule
\end{tabular}
\caption{\textbf{Ablation study on different loss functions.}} 
\label{tab:loss}
\end{table}

\subsection{Ablation Study}
\label{sec:ablation}
For all experiments under the ablation study, we utilize the Facade scene unless otherwise specified.

\subsubsection{Loss functions.}
In \cref{tab:loss}, we conduct an ablation study examining various combinations of loss functions using a 3D semantic occupancy prediction benchmark.
The weighted cross-entropy loss $\mathcal{L}_{\text{wce}}$ significantly enhances segmentation quality for less dominant classes, such as an advertisement board in the Others class.
Direct optimization of the Jaccard index using $\mathcal{L}_{\text{lovasz}}$ or incorporating $\mathcal{L}_{\text{affinity}}$ proves to be effective in refining the segmentation quality for the Pedestrian and Others classes, with a trade-off between the two classes.
Additionally, we find that optimizing the geometric and semantic aspects separately requires more time to converge.
Ultimately, the best performance is achieved by combining all of the aforementioned loss functions.

\subsubsection{Semantic scene understanding.}
We conduct an ablation study to assess the impact of semantic scene understanding (SSU) on 2D occupancy prediction and 3D instance occupancy prediction benchmarks.
By removing the Background class and retaining only the voxel occupancy state -- indicating whether a voxel is occupied (i.e., containing a Pedestrian) or Free -- the results presented in \cref{tab:ssu} demonstrate that integrating SSU significantly enhances performance in both tasks.
This experiment highlights the critical importance of contextual scene understanding, particularly in cross-scene scenarios where conditions vary.

\subsubsection{Pedestrian instance grouping (PIG).}
PIG is crucial for our model to predict instance and panoptic occupancy accurately.
It also improves qualitative results for the prediction of semantic occupancy, as visualized in \cref{fig:same_scene}, where it eliminates floating voxels in the absence of pedestrian instances.
However, these improvements are not reflected in the quantitative scores (\cref{tab:ablation_pig}) under the same-scene evaluation scheme, due to the large number of voxels compared to the number of uncertain voxels.
Even so, the synthetic-to-real results indicate a slight improvement in the Free class, where noises in the free space are significantly reduced.

\begin{table}[t] \centering
\setlength{\tabcolsep}{1.9mm}
    \begin{tabular}{*{10}{c}}
    \toprule
    \multicolumn{1}{c}{\multirow{2}{*}{SSU}} & \multicolumn{3}{c}{2D Pedestrian Occ.} & \multicolumn{3}{c}{3D Instance Occ.}\\ 
    \cmidrule(lr){2-4} \cmidrule(lr){5-7} 
    \multicolumn{1}{c}{} & MODA & MODP & F1  & AP & AP\textsubscript{25} & AP\textsubscript{50}\\
    \midrule
    -      &    66.5 &    77.1 &    81.6 & 38.2 & 66.2 & 39.0 \\
    \cmark & \B 75.4 & \B 80.3 & \B 87.5 & \B 40.2 & \B 70.0 & \B 39.8 \\
  \bottomrule
  \end{tabular}
    \caption{\textbf{Impact of semantic scene understanding (SSU) under synthetic-to-real evaluation on WildTrack.}} 
    \label{tab:ssu}
\end{table}

\begin{table}[t] \centering
\setlength{\tabcolsep}{1.4mm}
    \begin{tabular}{*{10}{c}}
    \toprule
    \multicolumn{1}{c}{\multirow{2}{*}{PIG}} & \small{Testing} & \multicolumn{2}{c}{Park} & \multicolumn{3}{c}{WildTrack}\\ 
    \cmidrule(lr){2-2} \cmidrule(lr){3-4} \cmidrule(lr){5-7} 
    \multicolumn{1}{c}{} & \multicolumn{1}{c}{\small{Training}} & mIoU & IoU\textsubscript{Ped.} & mIoU & IoU\textsubscript{Ped.} & IoU\textsubscript{Free} \\
    \midrule
    -      & Park & \B 91.8 & \B 62.9 &    75.1 & \B 47.9 & 87.8 \\
    \cmark & Park & \B 91.8 & \B 62.9 & \B 75.2 & \B 47.9 & \B 87.9 \\
  \bottomrule
  \end{tabular}
    \caption{\textbf{Impact of pedestrian instance grouping (PIG) on 3D semantic occupancy prediction.}} 
    \label{tab:ablation_pig}
\end{table}

\section{Conclusions}
This paper proposes a pioneering synthetic dataset specifically crafted for dense pedestrian scenarios in expansive multi-view environments.
Our dataset is the first of its kind, offering extensive 2D and 3D annotations that are particularly suited for urban surveillance contexts, thereby creating new opportunities for research in this area.
Through meticulous analysis, we present a robust baseline model that effectively addresses the challenges of multi-view pedestrian occupancy prediction.
Our findings underscore the importance of occupancy prediction and semantic scene understanding in boosting pedestrian detection performance, enabling our model to achieve state-of-the-art accuracy.
Furthermore, our model accurately reconstructs scenes with dynamic pedestrians when provided with 3D ground-truth data.
Importantly, we demonstrate excellent synthetic-to-real occupancy prediction performance through scene replication, paving the way for applications in real-world scenarios.

\section{Acknowledgments}
This work was partly supported by Institute of Information \& communications Technology Planning \& Evaluation (IITP) grant funded by the Korea government(MSIT)(RS-2023-00227592, Development of 3D Object Identification Technology Robust to Viewpoint Changes, 50\%), the Korea Institute of Police Technology (KIPoT) funded by the Korean National Police Agency \& Ministry of the Interior and Safety (RS-2024-00405100, 25\%), and the Korea Institute of Science and Technology (KIST) Institutional Program (Project No. 2E33001, 25\%).

\clearpage
\newpage

\appendix

\section{Supplementary Material}

\begin{figure*}[t] \centering
    \includegraphics[width=0.9\linewidth]{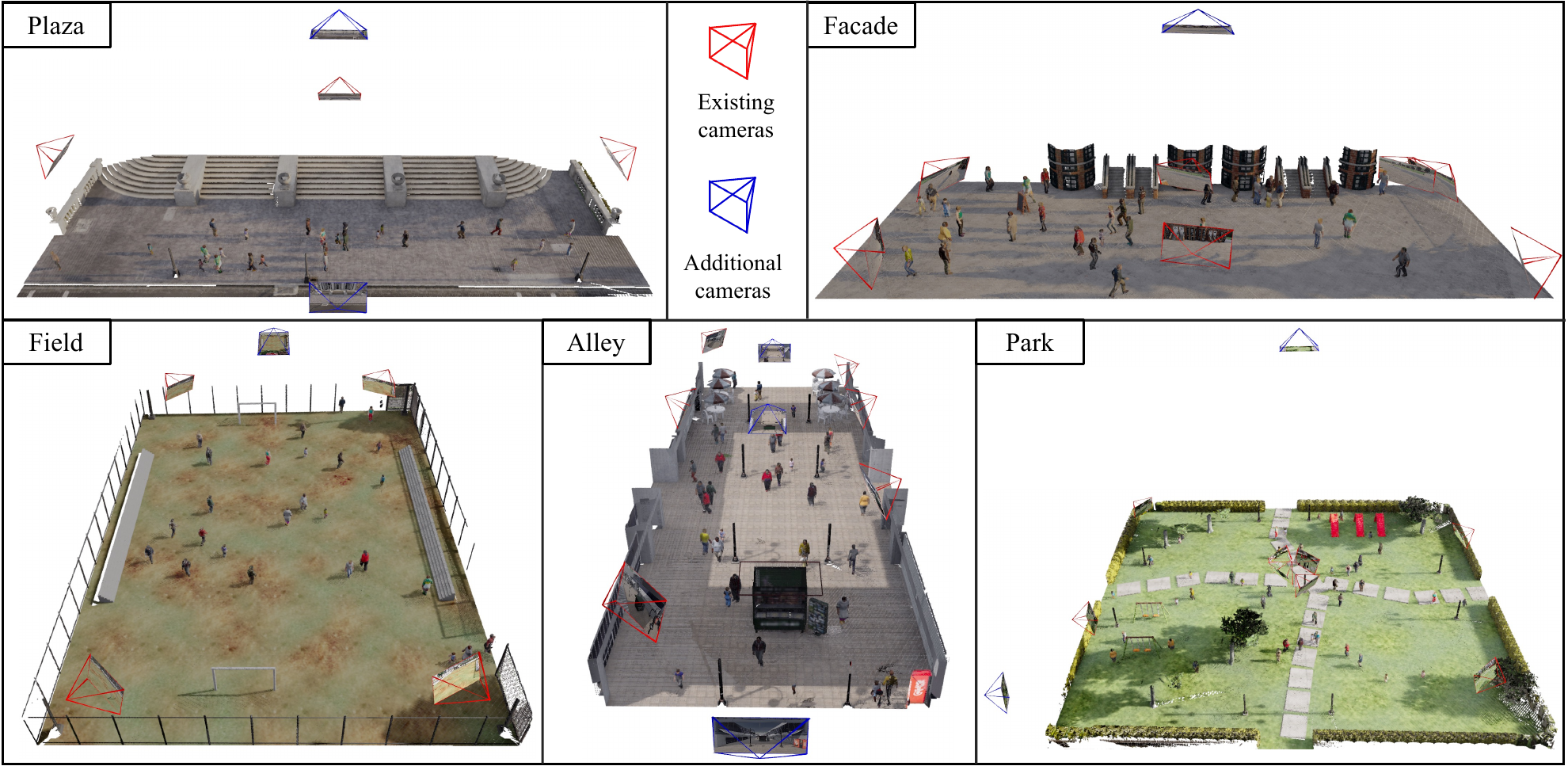}
    \caption{
    \textbf{Camera views used for generating a scene point cloud.}
    The number of additional camera views varies depending on the scene's characteristics, such as its size and level of occlusion.
    However, every scene includes an overhead camera to ensure comprehensive coverage.
    }
    \label{fig:scanning}
\end{figure*}

This supplementary material provides further details about occupancy label generation, evaluation metrics, and synthetic-to-real evaluation, followed by more experiments and visualizations.

\subsection{Occupancy Labels Generation for MVP-Occ}
Ideally, the occupancy labels are derived by voxelizing meshes of a static scene and moving pedestrians.
However, CARLA \cite{carla} does not support mesh extraction during simulation, and LiDAR scanners may fail to capture dense point cloud data.
To overcome this limitation, we generated the scene coordinates $\mathbf{P}_{\text{scene}}$ by fusing point clouds obtained from monocular views.
Let $\bar{u}_{ij}$ represents a 2D position $j$ in view $i$, which can be converted to a corresponding 3D coordinate in camera space, denoted as $x_{ij}$, using the ground-truth depth value $z_{ij}$:

\begin{equation}
    x_{ij} = z_{ij} \cdot K_i^{-1}\, \bar{u}_{ij}^T,
\end{equation}
where $K_i^{-1}$ is the inverse of intrinsic matrix for view $i$.

The lifted 3D coordinate $x_{ij}$ in camera space can be transformed to a scene coordinate $y_{ij}$ as:

\begin{equation}
    y_{ij} = R_i^{-1}\, x_{ij}^T,
\end{equation}
where $R_i^{-1}$ denotes the camera-to-world transformation matrix. 
$\mathbf{P}_{\text{scene}}$ is obtained by aggregating the scene coordinates from all views:

\begin{equation}
    \mathbf{P}_{\text{scene}} = \bigcup_{i \in N}\bigcup_{j \in HW} y_{ij},
\end{equation}
where $HW$ denotes the number of pixels in each image and $N$ represents the number of views.
We combine the scene coordinates from both existing and additional camera views, as depicted in \cref{fig:scanning} to handle occlusion.
The resulting $\mathbf{P}_{\text{scene}}$ is cropped within the area of interest and subsequently voxelized to create the occupancy labels, which can be represented in various formats, including RGB, semantic, or instance annotations, depending on the required task.

\begin{figure*}[t] \centering
    \includegraphics[width=\linewidth]{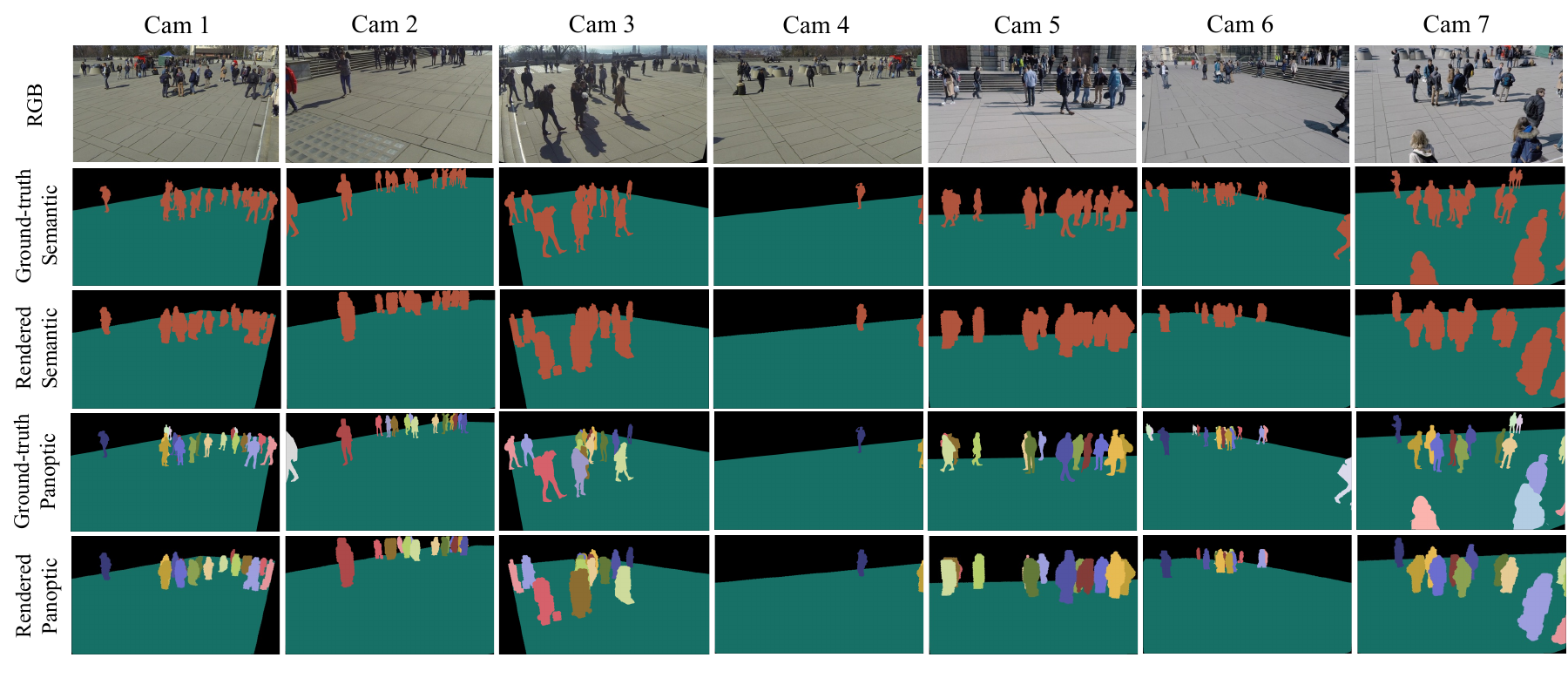}
    \caption{
    \textbf{Comparison between ground-truth and rendered segmentation masks.}
    Multi-view consistent rendered segmentation data are generated from our model trained on the Facade scene (third and last rows).
    }
    \label{fig:render}
\end{figure*}

\subsection{Details about Evaluation Metrics}

\subsubsection{2D pedestrian occupancy metrics.}
We employ commonly used metrics in multi-view detection methods \cite{mvdet, shot}, which include $\mathrm{MODA}$ (Multiple Object Detection Accuracy), $\mathrm{MODP}$ (Multiple Object Detection Precision), and $\mathrm{F1}$ scores.
Each metric evaluates different aspects of the model's performance in detecting pedestrians within the scene. 

\begin{equation}
\label{eq:moda}
    \mathrm{MODA} = 1 - \frac{\mathrm{FP} + \mathrm{FN}}{\mathrm{TP}+\mathrm{FN}}, 
\end{equation}

\begin{equation}
\label{eq:modp}
    \mathrm{MODP} = \frac{\sum 1 - d[d<t]/t}{\mathrm{TP}},
\end{equation}
where $\mathrm{TP}$, $\mathrm{FP}$ and $\mathrm{FN}$ denote the number of true positives, false positive, and false negative detections respectively.
If the distance $d$ between a detection and its ground truth is less than a threshold $t$ (set to \SI{50}{\centi\metre}), it is considered a true positive.
$\mathrm{MODA}$ score indicates the localization accuracy of pedestrians, whereas $\mathrm{MODP}$ score evaluates the localization precision of each detection, reflecting how closely the detected location is to the ground truth.

\begin{equation}
\label{eq:precision}
    \mathrm{Precision} = \frac{\mathrm{TP}}{\mathrm{TP}+\mathrm{FP}},
\end{equation}

\begin{equation}
\label{eq:recall}
    \mathrm{Recall} = \frac{\mathrm{TP}}{\mathrm{TP}+\mathrm{FN}},
\end{equation}

\begin{equation}
\label{eq:f1}
    \mathrm{F1} = 2 \times \frac{\mathrm{Precision} \times \mathrm{Recall}}{\mathrm{Precision}+\mathrm{Recall}}.
\end{equation}

$\mathrm{Precision}$ measures the proportion of true positive detections among all detected instances, while $\mathrm{Recall}$ assesses the proportion of true positives among all actual instances.
The $\mathrm{F1}$ score provides a balance between the two, offering a single measure of the model's overall performance.

\subsubsection{3D semantic occupancy metrics.}
In line with previous works on semantic scene completion \cite{sscnet, monoscene} and 3D semantic occupancy prediction tasks \cite{tpvformer, occnet}, we report voxel-level Intersection over Union ($\mathrm{IoU}$) metric for each class, as well as the mean Intersection over Union ($\mathrm{mIoU}$) metric across all classes.

The $\mathrm{IoU}$ metric is defined as follows:

\begin{equation}
\label{eq:iou}
    \mathrm{IoU} = \frac{\mathrm{TP}}{\mathrm{TP}+\mathrm{FP}+\mathrm{FN}}.
\end{equation}
% where $\mathrm{TP}$, $\mathrm{FP}$, and $\mathrm{FN}$ represent the number of true positive, false positive, and false negative predictions, respectively.

The $\mathrm{mIoU}$ metric is computed as the average $\mathrm{IoU}$ across all classes:

\begin{equation}
\label{eq:miou}
    \mathrm{mIoU} = \frac{1}{C_s}\sum_{c=1}^{C_s} \frac{\mathrm{TP}_c}{\mathrm{TP}_c+\mathrm{FP}_c+\mathrm{FN}_c},
\end{equation}
where $C_s$ denotes the number of semantic classes, which in our case is set to $5$, corresponding to the Free, Pedestrian, Ground, Walls, and Others classes.

\subsubsection{3D instance occupancy metrics.}
We use COCO metrics \cite{microsoftcoco} to evaluate the segmentation quality of instance occupancy prediction, which considers pedestrians only.
The average precision ($\mathrm{AP}$) metric, which is defined over multiple $\mathrm{IoU}$ thresholds can be calculated as follows:

\begin{equation}
\label{eq:ap}
    \mathrm{AP} = \frac{1}{T}\sum_{t\in T} \frac{\mathrm{TP}_t}{\mathrm{TP}_t+\mathrm{FP}_t+\mathrm{FN}_t},
\end{equation}
where $t$ is a certain $\mathrm{IoU}$ threshold and we consider $T = \{0.50, ..., 0.95 \}$ thresholds with a step size of 0.05, providing a comprehensive measure of the model's performance across varying levels of segmentation accuracy.
However, we find that in 3D occupancy, the IoU between the ground truth and prediction consistently exceeds 0.5, thus, we only report the final $\mathrm{AP}$ score.

\subsubsection{3D panoptic occupancy metrics.}
We use the standard panoptic quality ($\mathrm{PQ}$) metric \cite{panopticapi}, considering one thing class (Pedestrian) and three stuff classes (Ground, Wall, Others) to evaluate panoptic occupancy for the entire scene.
To compute $\mathrm{PQ}$, we first calculate the segmentation quality ($\mathrm{SQ}$) of true positive matches as:

\begin{equation}
\label{eq:sq}
    \mathrm{SQ} = \frac{\sum_{(p, g)\in \mathrm{TP_{t=0.5}}}\mathrm{IoU}(p, g)}{|\mathrm{TP_{t=0.5}}|},
\end{equation}
where matching criterion requires an $\mathrm{IoU}$ threshold greater than 0.5 between each prediction $p$ and ground truth $g$.

The recognition quality ($\mathrm{RQ}$) measures the detection accuracy as in $\mathrm{AP}$ and can be computed as follows:

\begin{equation}
\label{eq:rq}
    \mathrm{RQ} = \frac{|\mathrm{TP_{t=0.5}}|}{|\mathrm{TP_{t=0.5}}|+\frac{1}{2}|\mathrm{FP_{t=0.5}}|+\frac{1}{2}|\mathrm{FN_{t=0.5}}|}.
\end{equation}

$\mathrm{PQ}$ is defined as the multiplication of $\mathrm{SQ}$ and $\mathrm{RQ}$.

\begin{equation}
\label{eq:pq}
    \mathrm{PQ} = \mathrm{SQ} \times \mathrm{RQ}.
\end{equation}

\subsection{Details about Synthetic-to-Real Evaluation}

\subsubsection{Generating ground-truth segmentation masks}
We add new annotations for the WildTrack dataset \cite{wildtrack} by segmenting individual pedestrians within the area of interset (AoI). 
Each pixel within the AoI is categorized as either pedestrian, ground, or unlabelled (free space), and pedestrians outside the AoI are excluded for consistency with the original annotations.
This labelling process yields multi-view 2D semantic, instance, and panoptic segmentation masks for evaluation.

\subsubsection{Rendering by voxel ray marching.}
Due to the challenges in acquiring high-quality 3D occupancy labels for real scenes, we adopted an alternative approach by gauging 3D occupancy prediction performance with 2D segmentation labels.
Directly projecting the occupancy labels onto the image plane causes ambiguities in selecting the nearest voxel for each pixel.
Thus, we employ voxel ray marching \cite{amanatides1987fast} for each view to generate 2D segmentation predictions.
Each ray is colored according to the voxel it first intersects, and if the ray does not hit any voxels (indicating it is in free space), the color is set to zero.
We march the rays only within the scene boundary for 50 steps.
The minimum hit distance is set to \SI{10}{\centi\metre} and the maximum trace distance is to be ascertained as free to \SI{50}{\metre}.

\begin{table}[t] \centering
\setlength{\tabcolsep}{0.9mm}
      \begin{tabular}{l|*{3}{c}|*{3}{c}|c}
        \toprule
        \multirow{2}{*}{\small{Resolution}} & \multicolumn{3}{c|}{\small{3D Sem. Occ.}} & \multicolumn{3}{c|}{\small{3D Pan. Occ.} }& \multirow{2}{*}{\small{Time}} \\
        \cmidrule(lr){2-4} \cmidrule(lr){5-7}
        & mIoU & IoU\textsubscript{Ped.} & IoU\textsubscript{Bg.} & PQ & SQ & RQ & (sec)\\
        \midrule
        \small{360$\times$640}$\dagger$ & \U{79.8} & 58.4 & \B 90.7 & 58.9 & 76.2 & 72.3 & \B 8\\ 
        \small{720$\times$1280}         & \U{79.8} & \U{58.6} & \B 90.7 & \U{60.0} & \U{76.7} & \U{73.6} & \U{30}\\  
        \small{1080$\times$1920}        & \B 79.9 & \B 58.8 & \B 90.7 & \B 60.9 & \B 77.3 & \B 74.4 & 58 \\ 
      \bottomrule
      \end{tabular}
  \caption{
  \textbf{Evaluation results with different rendering resolutions.}
  $\dagger$ denotes the resolution we used in the main paper.
  Time represents the rendering time in seconds per image.
  }
  \label{tab:ablation_image_size}
\end{table}

\subsubsection{Evaluation scheme.}
We utilize the same metrics as in \cref{eq:moda,eq:modp,eq:precision,eq:recall,eq:f1} to evaluate 2D pedestrian occupancy for synthetic-to-real scenarios, as the WildTrack dataset \cite{wildtrack} already provides pedestrian locations on the ground plane.
However, we adapt \cref{eq:iou,eq:miou,eq:ap,eq:sq,eq:rq,eq:pq} to a view-level segmentation context, averaging over $N$ views for 3D occupancy evaluation.
Specifically, for the WildTrack dataset, the number of semantic classes $C_s$ is reduced to 3, corresponding to the Free, Pedestrian, and Ground classes.
The overlap thresholds are also adjusted to $T = \{0.25, ..., 0.70 \}$ with the same step size of 0.05 because of severe occlusion in 2D views.

\begin{table}[t] \centering
\setlength{\tabcolsep}{0.7mm}
      \begin{tabular}{c|c|*{15}{c}}
        \toprule
        \multicolumn{2}{c|}{\small{Location Decoder}} & \multicolumn{2}{c}{\small{2D Ped. Occ.}} & \multicolumn{2}{c}{\small{3D Sem. Occ.}} & \multicolumn{2}{c}{\small{3D Pan. Occ.}}\\
        \cmidrule(lr){3-4} \cmidrule(lr){5-6} \cmidrule(lr){7-8}
        \small{Compress} & \small{Head} & MODA & F1 & mIoU & IoU\textsubscript{Ped.} & PQ & SQ \\
        \midrule
        \small{Mean}$\dagger$ & \small{Conv.}$\dagger$ & \B 93.7 & \B 96.8 & \B 93.7 & \B 69.5 & \B 95.9 & \B 97.0 \\  
        \small{Conv.}         & \small{Conv.}$\dagger$ & 93.1 & 96.5 & 93.5 & 69.0 & 95.3 & 96.5 \\ 
        \midrule
        \small{Mean}$\dagger$ & \small{Dilated}        & 93.1 & 96.5 & \U{93.6} & \U{69.3} & \U{95.5} & \U{96.6} \\ 
        \small{Mean}$\dagger$ & \small{LKR}            & \U{93.3} & \U{96.6} & \U{93.6} & 69.0 & \U{95.5} & \U{96.6} \\ 
      \bottomrule
      \end{tabular}
  \caption{
  \textbf{Ablation study on location decoder. }
  ``Conv." denotes a $1\times 1$ 2D convolutional layer.
  }
  \label{tab:loc_decoder_ablation}
\end{table}

\subsubsection{Comparison between ground truth and rendered segmentation masks.}
In \cref{fig:render}, we present a qualitative comparison between the ground-truth segmentation masks and the rendered segmentation masks generated by our OmniOcc model trained on the Facade scene.
The rendered masks exhibit some degree of over-segmentation, resulting from using a relatively large voxel size of \SI{10}{\centi\metre}.
Despite this, the overall shape of a pedestrian is preserved.
Note that pedestrians close to the camera but not within the scene's boundary may not appear in the rendered views.

\subsubsection{Comparison between different rendering resolutions.}
We compare evaluation results across different rendering resolutions in \cref{tab:ablation_image_size}.
While using a larger image resolution results in higher scores for both semantic and panoptic occupancy predictions, it significantly increases rendering time -- taking 3 to 7 times longer -- without notably improving segmentation mask quality, due to the large voxel size used in the process.
Consequently, we select a 360$\times$640 image resolution, as it strikes the optimal balance between rendering time and the quality of the final results.

\subsection{More Experiments}
All the experiments in this section are conducted with the Facade scene.
The final setting used in the OmniOcc model is marked with $\dagger$.

\subsubsection{Ablation study on location decoder.}
In \cref{tab:loc_decoder_ablation}, we present two ablation studies on the location decoder.
The first study focuses on the compression mechanism used to convert the feature volume into a BEV feature map.
We find that a simple averaging mechanism outperforms a learnable aggregation layer.
The second study examines the final head of the location decoder, comparing a single convolutional layer with more complex structures, such as three dilated convolutional layers from MVDet \cite{mvdet} and the large kernel refiner (LKR) module from MVFP \cite{mvfp}.
The results indicate that these more sophisticated modules do not yield any improvements, highlighting that additional complexity does not necessarily translate to better performance in this scenario.

\begin{table}[t] \centering
\setlength{\tabcolsep}{0.7mm}
      \begin{tabular}{l*{2}{c}*{2}{c}cc}
        \toprule
        \multirow{2}{*}{\small{\#blocks}} & \multicolumn{2}{c}{\small{2D Pedestrian Occ.}} & \multicolumn{2}{c}{\small{3D Semantic Occ.}} & \small{Params.} & \multirow{2}{*}{\small{FPS}} \\
        \cmidrule(lr){2-3} \cmidrule(lr){4-5}
        & MODA & F1  & mIoU & IoU\textsubscript{Ped.} & (M) &  \\
        \midrule
        1           & 92.0 & 95.9 & 93.6 & 68.4 & \B 12.0 & \B 20\\ 
        2           & 92.2 & 96.0 & 93.6 & 68.6 & 12.6 & 18\\  
        3$\dagger$  & \B 93.7& \B 96.8 & \U{93.7} & \U{69.5} & 13.1 & 17 \\ 
        4           & \U{93.5} & \U{96.7} & \B 93.9 & \B 70.3 & 13.7 & 16 \\ 
      \bottomrule
      \end{tabular}
  \caption{
  \textbf{Ablation study on voxel encoder.} 
  \#blocks: the number of residual blocks used in the voxel encoder.
  }
  \label{tab:ablation_voxel_encoder}
\end{table}

\subsubsection{Ablation study on voxel encoder.}
In \cref{tab:ablation_voxel_encoder}, we investigate how the number of residual blocks in the voxel encoder affects model performance.
The results reveal that increasing the number of residual blocks typically leads to better pedestrian detection and occupancy prediction accuracies.
However, while adding more than three blocks does improve 3D occupancy prediction, it results in a decline in 2D occupancy prediction performance.
Thus, we choose to use three residual blocks in the voxel encoder.

\begin{table}[t] \centering
    \setlength{\tabcolsep}{0.8mm}
    \begin{tabular}{l*{10}{c}}
        \toprule
        Method & Train & Test & MODA & F1 \\
        \midrule
        GMVD           & Facade & WildTrack & 46.3 & 71.9 \\
        MVFP           & Facade & WildTrack & 52.6 & 74.9 \\ 
        OmniOcc (ours) & Facade & WildTrack & \U{75.4}& \U{87.5}\\  
        \midrule
        OmniOcc (ours) & MVP-Occ & WildTrack & \B 78.8 & \B 89.7 \\
        \bottomrule
    \end{tabular}
    \caption{
    \textbf{Synthetic-to-real evaluation with training on all scenes from MVP-Occ.}
    Training the model on all scenes further increases the performance.} 
    \label{tab:all_scenes_train}
\end{table}

\subsubsection{Training on all scenes.}
In \cref{tab:all_scenes_train}, we show the synthetic-to-real performance with the model trained on all scenes from the proposed dataset.
The results indicate that training solely on the Facade scene, which closely resembles WildTrack, already provides strong performance of 87.5 F1 score in direct synthetic-to-real evaluation.
Conduting unified training across scenes enhances the performance to 89.7 F1 score though the improvement is minimal compared to the performance with strong scene replication.

\begin{table}[t] \centering
    \setlength{\tabcolsep}{0.8mm}
    \begin{tabular}{l*{10}{c}}
        \toprule
        Backbone & mIoU & IoU\textsubscript{Ped.} & IoU\textsubscript{Bg.} & Params & Mem. & FPS \\
        \midrule
        R18$\dagger$    & 93.7 & 69.5 & 99.8 & \B 13.1M & \B 4.6GB & \B 17 \\ 
        R50             & 93.9 & 70.3 & 99.8 & 25.6M & 5.3GB & 11\\  
        R101            & \B 94.3 & \B 71.9 & \B 99.9 & 44.5M & 5.4GB & 8 \\
        \bottomrule
    \end{tabular}
    \caption{
    \textbf{Impact of different backbone networks, and computational cost analysis.}
    FPS is measured on the Facade scene, which has 7 cameras with 720$\times$1280 image size on a single A100 GPU.} 
    \label{tab:backbone}
\end{table}

\subsubsection{Computational cost analysis.}
In \cref{tab:backbone}, we analyze the performance and latency analysis of employing different backbone networks within our model.
Larger backbones improve accuracy but slow down the model greatly.
Our model with ResNet-18 backbone achieves 17 FPS, experiencing only a minimal 0.8\% decrease in IoU score for pedestrians compared to the ResNet-50 backbone.
Notably, the model's speed and memory usage can vary depending on the voxel grid size and the number of cameras utilized.
In \cref{tab:backbone}, we used the Facade scene with a grid size of 360$\times$120$\times$30 and 7 cameras operating at a resolution of 720$\times$1280.

\begin{table}[t] \centering
    \setlength{\tabcolsep}{0.8mm}
    \begin{tabular}{l*{10}{c}}
        \toprule
        Method & Train & Test & MODA & F1 \\
        \midrule
        GMVD           & MVP-Occ & CityStreet & 34.8 & 51.8 \\
        MVFP           & MVP-Occ & CityStreet & 45.6 & 63.2 \\ 
        \midrule
        OmniOcc (ours) & MVP-Occ & CityStreet & \B 57.3 & \B 78.0 \\
        \bottomrule
    \end{tabular}
    \caption{
    \textbf{Synthetic-to-real evaluation with training on all scenes from MVP-Occ.}
    } 
    \label{tab:citystreet_results}
\end{table}

\subsubsection{Results on CityStreet.}
In \cref{tab:citystreet_results}, we report the results of synthetic-to-real evaluation on CityStreet dataset \cite{citystreet}, which has larger scene dimensions with more pedestrians compared to the WildTrack dataset \cite{wildtrack}.
All models are trained on the synthetic scenes from the proposed dataset and directly evaluated on the real CityStreet dataset.
The evaluation threshold $t$ is increased to \SI{2}{\metre} because of the larger scene size and nosiy ground-truth data.
Our model still outperforms compared methods while showing the diversity of our dataset in terms of generalization to this challenging scene.

\subsection{More Visualizations}
All visualizations under this section are generated with our OmniOcc model, apart from the ground-truth data.

\subsubsection{Scene variations.}
In \cref{fig:all2wt}, we depict the differences in scene conditions between our MVP-Occ dataset and the WildTrack dataset, as viewed through RGB images.
A key difference is the camera setup: in our dataset (except for the Facade scene), cameras are mounted at heights typical of CCTV installations, ranging from \SI{3} to \SI{8}{\metre}.
In contrast, the WildTrack dataset features cameras installed at an average human height of about \SI{2}{\metre}.
As a result, pedestrians in our scenes appear much smaller and farther from the camera, which complicates the synthetic-to-real evaluation on WildTrack. 
Additionally, the Alley and Park scenes in our dataset are cluttered with distractions and background objects, highlighting the necessity of robust scene understanding for accurate pedestrian recognition.

\subsubsection{Qualitative results on WildTrack.}
\begin{figure*}[t] \centering
    \includegraphics[width=0.95\linewidth]{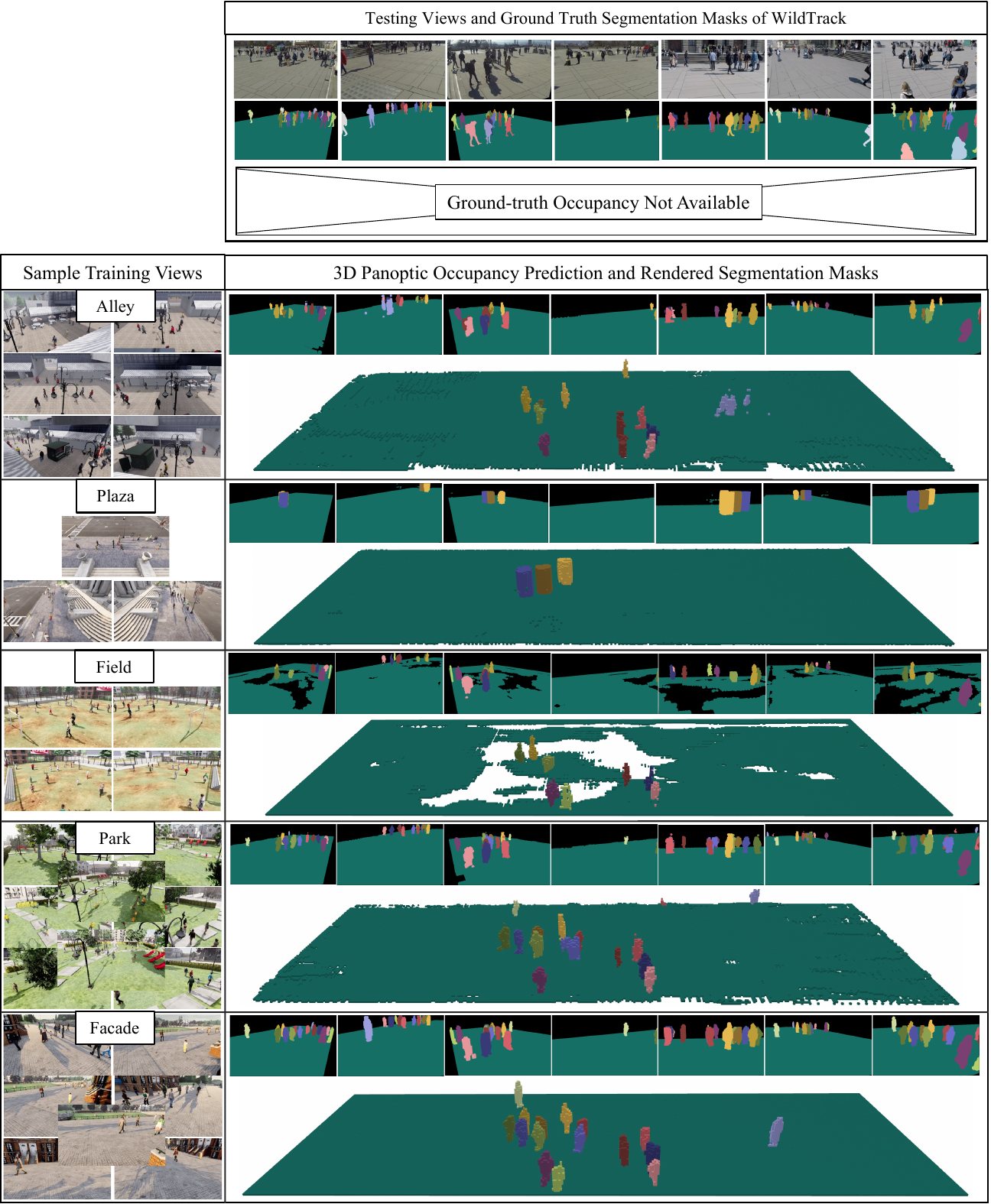}
    \caption{
    \textbf{Qualitative results of 3D panoptic occupancy prediction under synthetic-to-real evaluation on WildTrack.}
    The model is trained on each scene of MVP-Occ and tested on the WildTrack. 
    Note that there is no ground-truth 3D occupancy data available in the WildTrack scene.
    }
    \label{fig:all2wt}
\end{figure*}

In \cref{fig:all2wt}, we present the qualitative results of 3D panoptic occupancy prediction and rendered segmentation masks on WildTrack, generated by our model, which was trained on each scene of MVP-Occ.
Importantly, the WildTrack dataset does not have accompanying occupancy ground truth.
Among the MVP-Occ scenes, only the Alley, Plaza, and Facade scenes share structural and semantic similarities with WildTrack.
As a result, these scenes enable more accurate modeling of ground plane structures compared to the Field and Park scenes, where grass textures replace the tile patterns found in WildTrack.

As previously discussed, the Alley and Plaza scenes feature high-mounted cameras, which complicate generalization to WildTrack compared to the Facade scene.
This challenge is particularly evident in the Plaza scene, where the limited number of cameras (only three) makes pedestrians difficult to identify.
A similar issue occurs in the Field scene with just four cameras.

Conversely, the Park scene, with its large number of camera views and diverse pedestrian distribution (characterized by varied pedestrian density and movement patterns), aids the model in focusing more effectively on pedestrians, thereby enhancing detection and occupancy predictions.
The Facade scene produces the best results overall, achieving near-perfect predictions of ground plane structure and high-quality occupancy predictions across the board.

\subsubsection{Qualitative results on MVP-Occ.}
\begin{figure*}[t] \centering
    \includegraphics[width=\linewidth]{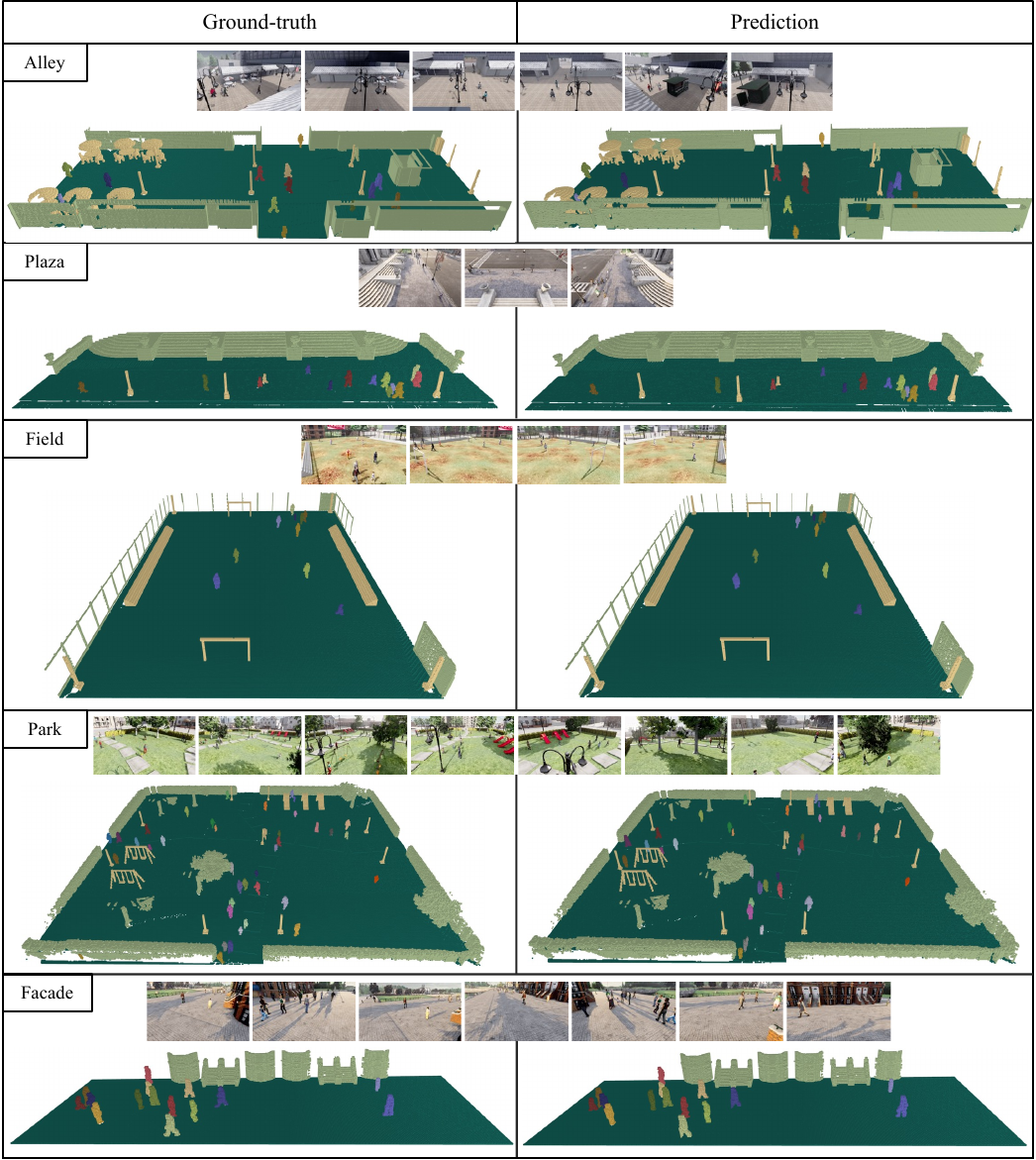}
    \caption{
    \textbf{Qualitative results of 3D panoptic occupancy prediction under same-scene evaluation on MVP-Occ.}
    The model is trained and tested on each scene but with different splits.
    }
    \label{fig:all_same_scene}
\end{figure*}

We present the qualitative results of 3D occupancy prediction on each scene of MVP-Occ in \cref{fig:all_same_scene}.
The model demonstrates strong performance in all scenes, achieving phenomenal occupancy predictions while accurately capturing pedestrian postures.

\subsection{Future Works}
Although our OmniOcc model has demonstrated outstanding performance, particularly in same-scene evaluations, there is still room for improvement in synthetic-to-real evaluations.
While the Facade scene yields excellent occupancy prediction results on the WildTrack dataset, the critical factors contributing to successful synthetic-to-real transfer need further investigation.
Identifying and understanding these factors could be a significant step forward, offering valuable insights for real-world applications.

Moreover, the challenge of acquiring accurate 3D ground-truth data in real-world environments has led to the exploration of alternative approaches, such as supplementing training with 2D supervision.
Techniques like neural rendering \cite{nerf}, as demonstrated in \cite{occnerf, renderocc, selfocc}, offer promising avenues in this regard.
However, accurately capturing small objects like pedestrians is yet to be addressed to enhance their effectiveness in our dense pedestrian scenarios.

\clearpage
\newpage

\bibliography{aaai25}

\end{document}